\documentclass[journal,twoside,web]{ieeecolor}
\usepackage{generic}
\usepackage[T1]{fontenc} 
\usepackage{cite}
\usepackage{amsmath,amssymb,amsfonts}
\usepackage{algorithmic}
\usepackage{graphicx}
\usepackage{algorithm,algorithmic}
\usepackage{hyperref}
\hypersetup{hidelinks=true}
\usepackage{textcomp}
\usepackage[table]{xcolor}
\usepackage{color,soul}
\usepackage[table]{xcolor}
\usepackage{tabularx}
\usepackage{subfig}
\colorlet{soulblue}{blue!20}
\usepackage{booktabs}
\usepackage{multirow}
\usepackage{caption}
\usepackage{subcaption}
\usepackage{placeins}

\def\BibTeX{{\rm B\kern-.05em{\sc i\kern-.025em b}\kern-.08em
    T\kern-.1667em\lower.7ex\hbox{E}\kern-.125emX}}
\begin{document}
\title{Sedentary Behavior Classification for Wearable Sensors  with a CNN-BiLSTM Model}
\author{Yuliang Chen, Weiwei Shi, Rong Zablocki, Animesh Kumar, Jordan A. Carlson, Sheri J. Hartman, Mikael Anne Greenwood-Hickman, Paul R. Hibbing,  Marta Jankowska, Jay Yang, Jingjing Zou$^*$, Arun Kumar$^*$, and Loki Natarajan$^*$
\thanks{This work was partially supported by funding from the National Institute of Diabetes and Digestive and Kidney Diseases (R01DK114945) and the National Heart, Lung, and Blood Institute (1R01HL168535, R01HL130483, R01HL166802).}
\thanks{Yuliang Chen is with the Hal{\i}c{\i}o\u{g}lu Data Science Institute, University of California, San Diego, La Jolla, CA 92093, USA (Email: yuc204@ucsd.edu).}%
\thanks{Loki  Natarajan,  Jingjing  Zou,  Rong  Zablocki,  Sheri  J.  Hartman,  and  Weiwei  Shi,  are  with  the  Herbert  Wertheim School  of  Public  Health  and  Human  Longevity  Science, University of California, San Diego, La Jolla, CA 92093, USA (Email:   lnatarajan@ucsd.edu;   j2zou@health.ucsd.edu; rzablocki@health.ucsd.edu; sjhartman@health.ucsd.edu; w2shi@ucsd.edu).}%
\thanks{Animesh Kumar is with the Department of Computer Science and Engineering, University of California, San Diego, La Jolla, CA 92093, USA (Email: ank028@ucsd.edu).}%
\thanks{Jordan A. Carlson is with the Center for Children’s Healthy Lifestyles \& Nutrition, Children's Mercy Kansas City, University of Missouri-Kansas City, Kansas City, MO 64108, USA. (Email: jacarlson@cmh.edu).}%
\thanks{Mikael Anne Greenwood-Hickman is with the Kaiser Permanente Washington Health Research Institute, Seattle, WA 98101, USA (Email Mikael.Anne.Greenwood-Hickman@kp.org).}%
\thanks{Paul R. Hibbing is with the Department of Kinesiology and Nutrition, University of Illinois Chicago, Chicago, IL, 60612 USA  (Email: paulhibbing@gmail.com)}%
\thanks{Marta Jankowska and Jiue-An Yang are with the Beckman Research Institute, City of Hope Cancer Center, Department of Population Sciences, Duarte, CA 91010 (Email: mjankowska@coh.org; jyang@coh.org).}%
\thanks{Arun Kumar is with the Department of Computer Science and Engineering and the Hal{\i}c{\i}o\u{g}lu Data Science Institute, University of California, San Diego, La Jolla, CA 92093, USA (Email: akk018@ucsd.edu).}%
\thanks{*Jingjing Zou, Arun Kumar and Loki Natarajan are co-senior authors}
}

\maketitle

\begin{abstract}
Accurate detection of sedentary behavior is important for studying health risks related to prolonged sitting, but posture-based classification remains challenging with wearable sensors, especially at the wrist. We study whether a deep learning model trained on hip-worn accelerometer data can transfer to wrist-worn accelerometer data for sitting versus non-sitting classification. We use CHAP, a CNN-BiLSTM model originally developed for hip accelerometers, and evaluate its zero-shot performance on wrist data as well as its adaptation through finetuning with varying amounts of labeled wrist data. Experiments are conducted on the iWatch dataset with ground-truth posture labels derived from wearable cameras. The hip-trained model performs strongly on hip data without retraining, but accuracy drops on wrist data due to sensor placement shift. Finetuning CHAP provides consistent advantages over transformer models trained from scratch. These findings suggest that hip-based pretraining provides a useful starting point for wrist deployment, while highlighting the need for wrist-specific adaptation to handle higher signal variability.
\end{abstract}


\section{Introduction}\label{intro}
Sedentary behavior is a recognized risk factor for chronic disease \cite{Biswas2015}. Total sedentary time and prolonged sitting bouts are associated with adverse cardiometabolic profiles and increased morbidity and mortality risk, even after accounting for moderate-to-vigorous physical activity (\cite{Dunstan2021} and references therein). Frequent breaks from sitting, especially those that break up long sitting bouts, are associated with lower metabolic risk \cite{Healy2008}. These clinical links motivate accurate measurement of sedentary bout duration and breaks in sedentary time in free-living settings.

Wearable accelerometers are widely used to quantify physical activity and inactivity because they enable objective, continuous monitoring outside the laboratory and reduce biases inherent in self-report. In large behavioral and population studies, hip or wrist accelerometers are practical for long-term wear and can be deployed at scale. However, most studies quantify sedentary behavior using energy-expenditure-based cut points applied to hip-worn or wrist-worn acceleration signals \cite{GGIR}. This intensity-only approach does not explicitly capture posture, a defining component of sedentary behavior. As a result, low-movement activity (i.e., standing) can be misclassified as sedentary, and transitions can be miscounted, leading to overestimation of breaks in sedentary time and underestimation of sedentary bout durations \cite{Bellettiere2021Agreement}. Early wrist-based efforts applied traditional machine learning methods such as support vector machines with hand-crafted features \cite{Chowdhury2018,Mannini2013}, but their reliance on manual feature engineering limited their capacity to model longer temporal structure. Recent work in human activity recognition has increasingly adopted deep learning models that learn representations directly from raw accelerometer signals along three axes \cite{Chap1.0}. To better capture posture, the CHAP model \cite{Chap1.0} was designed for hip-worn accelerometers, where the signal more directly reflects trunk orientation and whole-body posture. When compared to a gold-standard thigh-worn device, the CHAP model achieves over 90\% balanced accuracy for sitting versus non-sitting classification using triaxial hip-worn accelerometers.

The open question is whether this hip-trained model can be used in wrist-only settings without collecting new labels, since wrist-worn devices are already common in consumer and research wearables, are easier to wear continuously, and often yield higher compliance in long-term monitoring. At the same time, cross-body position transfer is underexplored because wrist motion is more task-dependent and the sensor orientation and signal statistics differ from the hip, even when the underlying behavior is the same \cite{Karas2019Accelerometry}. Evidence that activity recognition models can generalize across related domains, for example from adults to children \cite{AdultChileTransfer}, suggests that cross-device generalization may also be possible, motivating a systematic test of whether hip-to-wrist transfer can work and how much labeled wrist data is needed when it does not.

In this work, we test cross-placement generalization by applying the hip-trained CHAP model to wrist accelerometer data with no retraining. We then quantify adaptation by finetuning CHAP with increasing amounts of labeled wrist data and tracking how performance changes as more labels are added. In parallel, we train transformer-based models from scratch on the same dataset to compare against transfer under matched data conditions and to provide wrist-specific baselines.
\section{Methods}\label{methods}
\begin{figure}
    \centering
    \includegraphics[width=\linewidth,trim=0pt 0pt 18pt 0pt,clip]{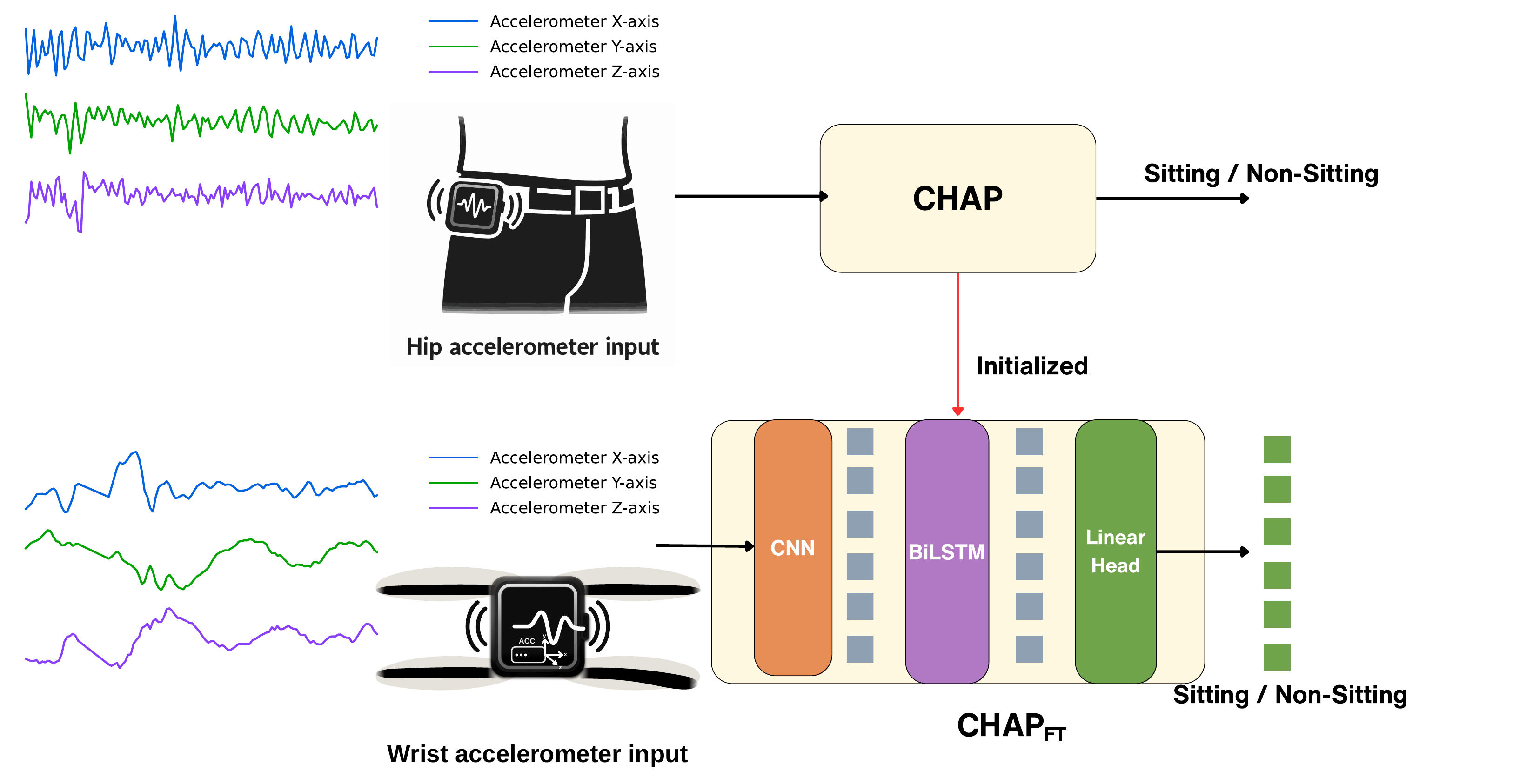}
    \caption{\textbf{Overview of the study}. We use CHAP weights pretrained on hip accelerometer data from a prior study, then finetune them on iWatch wrist / hip accelerometer data for the same sitting versus non-sitting classification task.}
    \label{fig:chap2}
\end{figure}
\subsection{Data} \label{data}
\textbf{Study Sample and Data Collection.} 
Our study sample comprised adult participants in the iWatch study, an observational study that objectively measured sitting and physical activity patterns using wearable sensors. The study enrolled 148 adults aged 24 to 85 years, of whom 50\% were female, 22\% were non-White, and 20\% were Hispanic. The study was approved by the Institutional Review Board (IRB) at the University of California, San Diego, IRB\# 111160, date 07/12/2011.  All participants provided their informed consent.

\textbf{Data Collection and Processing.} 
The study used a person-worn camera, SenseCam (Vicon Revue), which captured first-person images approximately every 20 seconds. Participants wore the camera on a lanyard around their necks for 7 days (excluding nighttime, when devices were charged). SenseCam images were imported into the Clarity SenseCam Browser \cite{Doherty2011SenseCam}; a standardized annotation protocol was developed and inter-rater reliability was established \cite{Kerr2013SenseCamSedentary}. Annotated images classified participant activity into the following categories: {\em{Sedentary/Sitting, Standing Still, Standing Moving, Walking/Running, Vehicle, Nonwear, Uncodable.}} For this study, these categories were collapsed into two binary classes: Sitting ({\em{Sedentary, Vehicle}}) and Nonsitting ({\em{Standing Still, Standing Moving, Walking/Running}}); nonwear and uncodable intervals were excluded.

Data collection also included hip- and wrist-worn ActiGraph GT3X+ accelerometers \cite{Cain2013AccelerometersYouth,Matthews2012BestPracticesMonitors}, attached to a belt (hip/waist) or a wristband (non-dominant wrist). Participants typically wore the monitors for 7 days during waking hours (hip) or 24 hours (wrist). The devices recorded raw triaxial acceleration (x, y, z axes) in gravitational units at a 30 Hz sampling rate.

Following the protocol described in \cite{supunCNN}, data from the GT3X+ and SenseCam devices were temporally aligned and merged to generate a single 30 Hz output file per participant. A few participants lacked concurrent wear of both devices, resulting in a final analytic sample of 141 participants for the hip sensor and 143 for the wrist sensor. Non-wear periods were identified by the Choi algorithm \cite{choi} based on consecutive zero counts of vector magnitude ($=\sqrt{x^2+y^2+z^2}$) from GT3X+ minute-level count files and subsequently removed. The resulting files were segmented into non-overlapping 10-s windows, yielding 2,456,142 windows (6,822 h) for hip-worn data and 2,960,580 windows (8,223 h) for wrist-worn data. The dataset was randomly partitioned into training, validation, and test sets using a 60-15-25 split. Partitioning was performed at the participant level, ensuring that data from the same individual did not appear across subsets, thereby avoiding data leakage and supporting evaluation on unseen users.

\textbf{Label Distribution.} The dataset is highly imbalanced, with 1,176,336 sitting versus 403,032 non-sitting windows for the hip sensor, and 1,189,272 sitting versus 427,896 non-sitting windows for the wrist sensor. This skew can bias models toward predicting sitting more often, making balanced accuracy a more appropriate metric than overall accuracy. To address the imbalance, we used a weighted binary cross-entropy loss that upweights the minority non-sitting class (details in Section~\ref{implementation}).

\subsection{CHAP Model}
\textbf{Model Purpose and Training Context.} CHAP \cite{Chap1.0} is a Convolutional Neural Network--Bidirectional Long Short-Term Memory (CNN-BiLSTM) sequence model developed for sitting versus non-sitting classification from hip-worn ActiGraph GT3X+ accelerometer data, trained on 1,397 adults using labels derived from concurrently worn thigh-mounted activPAL devices. Here, we test whether its architecture and pretrained weights transfer to the iWatch dataset, which includes both hip and wrist placements.



\textbf{Model Architecture.} As shown in Figure~\ref{fig:chap2}, each prediction uses a 7-minute triaxial accelerometer segment resampled at 10 Hz. The segment is partitioned into $T=42$ consecutive, non-overlapping 10-second windows, yielding an input tensor $X \in \mathbb{R}^{T \times 100 \times 3}$, where the $t$-th window is $X_t \in \mathbb{R}^{100 \times 3}$. A CNN encodes each window independently into a 128-dimensional feature vector, $h_t = f_{\mathrm{CNN}}(X_t) \in \mathbb{R}^{128}$. A BiLSTM then captures temporal dependencies across the full 7-minute context, producing context-aware features $\tilde{h}_t = f_{\mathrm{BiLSTM}}(h_{1:T})_t$. A shared linear classifier maps each representation to a scalar score $\ell_t = w^\top \tilde{h}_t + b$, and the sitting probability is $p_t = \sigma(\ell_t)$, where $\sigma(\cdot)$ denotes the sigmoid function.

\textbf{Transferability.} We test whether a CHAP model pretrained on hip data generalizes to a new cohort and a different device placement by evaluating two settings on the iWatch hip and wrist datasets. \textbf{(1) CHAP Zero-shot (CHAP\textsubscript{ZS})} applies the hip-pretrained model to iWatch data with no parameter updates. \textbf{(2) CHAP Finetuning (CHAP\textsubscript{FT})} updates all model parameters on the iWatch training set and is evaluated on the corresponding iWatch hip or wrist test set. Comparing CHAP\textsubscript{ZS} and CHAP\textsubscript{FT} separates direct transfer from gains due to supervised adaptation.

\subsection{Transformer-based Models}
Given the strong performance of Transformer models \cite{Transformer} and their increasing adoption for sensor time series modeling \cite{sundial,patchtst}, we include a standard Transformer-based model as a baseline to compare to CHAP\textsubscript{FT}. Specifically, we tokenize the multichannel accelerometer signal into fixed length temporal patches and embed each patch as a token, following the generic patch embedding design popularized by ViT and PatcTST \cite{ViT,patchtst}. We train the \textbf{ViT\textsubscript{Small}} model on the iWatch training set with supervised training and evaluate on the test set. 



\section{Results}\label{results}
We first examine the hip--wrist distribution gap, then evaluate the three model settings, CHAP\textsubscript{ZS}, CHAP\textsubscript{FT}, and ViT\textsubscript{Small},  on the iWatch test set using SenseCam-derived labels as ground truth. Balanced accuracy is the primary metric; F1 score is supplementary. We also report subject-level performance, since the subject is typically the unit of analysis in clinical studies.

\subsection{Hip-Wrist Distribution Gap} 
\label{sec:distribution}
\begin{figure}
    \centering
    \includegraphics[width=1\linewidth]{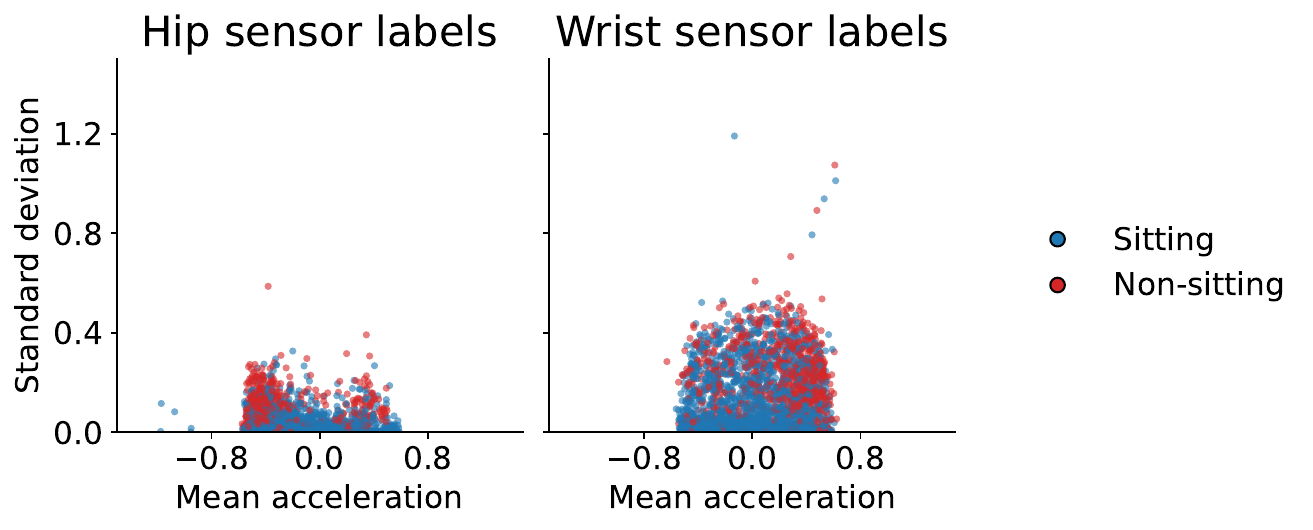}
    \caption{\textbf{Scatter plots of mean acceleration versus standard deviation (gravitational units) within 10-sec windows for hip and wrist sensors, colored by activity labels}. Blue denotes sitting and red denotes non-sitting.}
    \label{fig:joint_hexbin_hip_wrist}
\end{figure}

Figure \ref{fig:joint_hexbin_hip_wrist} compares the joint distributions of window-level mean and standard deviation of acceleration for hip and wrist sensors using 10-second windows, with sitting in blue and non-sitting in red. For each window, we compute the per-axis mean and standard deviation and then average across axes. In both placements, sitting and non-sitting overlap strongly with no clear boundary. The hip distribution is compact and concentrated at low variance, while the wrist distribution is more spread due to incidental arm motion, including during sitting. This indicates that simple statistics cannot reliably separate the two classes, motivating the use of nonlinear models that exploit richer temporal structure.

The figure also shows a clear shift between hip and wrist distributions. We quantify this via the Jensen--Shannon Distance (JSD)~\cite{JSD} between the two joint histograms; the JSD = 0.486, indicating a large distribution gap. This gap suggests that transferring a hip-trained model to wrist data without adaptation will likely reduce performance and that wrist-specific finetuning is needed to handle the placement shift.

\subsection{Input-level Performance}
\label{model_accuracy}



Table~\ref{tab:submit-result} summarizes cross-placement transfer learning results.
CHAP\textsubscript{ZS} achieves 88.74\% test balanced accuracy on the hip dataset, and finetuning on iWatch hip yields minimal improvement, indicating that the hip-pretrained representation transfers well to hip data in this cohort. On the wrist dataset, CHAP\textsubscript{ZS} drops to 71.83\%, but CHAP\textsubscript{FT} raises performance to 82.56\%. Combined with the hip-to-wrist distribution shift in Figure~\ref{fig:joint_hexbin_hip_wrist}, this indicates that the learned representations are not fully placement-agnostic and require wrist-specific adaptation for reliable transfer.

A randomly initialized ViT\textsubscript{Small} trained on iWatch underperforms CHAP\textsubscript{FT} despite having more parameters. CHAP\textsubscript{FT} benefits from initialization already trained for sitting vs.\ non-sitting on hip data, so finetuning mainly adjusts the model to the iWatch cohort and, for wrist, to the placement shift. In contrast, ViT\textsubscript{Small} must learn both useful representations and the decision boundary from scratch using only iWatch labels.

\begin{table}[t]
\centering
\caption{Balanced accuracy (Bal acc) and F1-score (\%) of CHAP\textsubscript{ZS}, CHAP\textsubscript{FT}, and ViT\textsubscript{Small} across hip and wrist placements.}
\label{tab:submit-result}
\setlength\tabcolsep{6pt}
\resizebox{\linewidth}{!}{%
\begin{tabular}{llccc ccc ccc}
\toprule
\textbf{Placement} & \textbf{Metric} 
& \multicolumn{3}{c}{CHAP\textsubscript{ZS}} 
& \multicolumn{3}{c}{CHAP\textsubscript{FT}} 
& \multicolumn{3}{c}{ViT\textsubscript{Small}} \\
\cmidrule(lr){3-5}\cmidrule(lr){6-8}\cmidrule(lr){9-11}
& & \textbf{Train} & \textbf{Validation} & \textbf{Test}
  & \textbf{Train} & \textbf{Validation} & \textbf{Test}
  & \textbf{Train} & \textbf{Validation} & \textbf{Test} \\
\midrule

\multirow{2}{*}{Hip} 
& Bal acc & 90.43 & 92.34 & \textbf{88.74} & 90.36 & 92.29 & 88.71 & 86.62 & 87.41  & 85.22 \\
& F1      & 90.95 & 92.66 & 89.72 & 91.28 & 92.99 & \textbf{90.04} & 88.85 & 89.84  & 87.77 \\
\midrule

\multirow{2}{*}{Wrist} 
& Bal acc & 73.79 & 73.19 & 71.83 & 88.71 & 86.01 & \textbf{82.56} & 82.84 & 82.20 & 79.39 \\
& F1      & 69.83 & 69.72 & 69.40 & 89.26 & 88.02 & \textbf{84.97} & 86.27 & 86.38 & 83.45 \\
\bottomrule
\end{tabular}%
}
\end{table}

Table~\ref{tab:confusion_matrix_side_by_side} compares confusion matrices across placements and models.
The results reveal a wrist-specific, asymmetric error pattern: CHAP\textsubscript{FT} labels 22.9\% of non-sitting as sitting versus 12.0\% of sitting as non-sitting. This likely reflects low-motion upright activities (i.e., standing) that produce wrist signals similar to sitting, inflating sitting predictions. The hip classifier shows fewer and more balanced errors (14.1\% non-sitting$\rightarrow$sitting; 8.5\% sitting$\rightarrow$non-sitting), consistent with hip motion being more directly coupled to posture. ViT\textsubscript{Small} amplifies the same wrist bias (30.7\% non-sitting$\rightarrow$sitting) and slightly overpredicts sitting prevalence (72.3\% vs.\ 70.8\%), while CHAP\textsubscript{FT} remains closer to the true rate (69.0\%). Overall, CHAP\textsubscript{FT} achieves higher balanced accuracy by improving non-sitting recall on wrist (77.1\% vs.\ 69.3\%), indicating more reliable separation of sitting from low-motion non-sitting.

\begin{table}[t]
\centering
\setlength{\tabcolsep}{1pt}
\renewcommand{\arraystretch}{0.95}
\begin{minipage}[t]{0.45\linewidth}
\centering
\subfloat[CHAP\textsubscript{FT}, Hip]{
\resizebox{0.7\linewidth}{!}{%
\begin{tabular}{l|cc}
 & \multicolumn{2}{c}{\textbf{Predicted}} \\
\cmidrule(lr){2-3}
\textbf{Actual} \quad & \textbf{Non-sit} & \textbf{Sit} \\
\midrule
Non-sit & 0.86 & 0.14 \\
Sit & 0.09 & 0.91 \\
\bottomrule
\end{tabular}}
}
\end{minipage}
\hfill
\begin{minipage}[t]{0.45\linewidth}
\centering
\subfloat[CHAP\textsubscript{FT}, Wrist]{
\resizebox{0.7\linewidth}{!}{%
\begin{tabular}{l|cc}
 & \multicolumn{2}{c}{\textbf{Predicted}} \\
\cmidrule(lr){2-3}
\textbf{Actual} \quad & \textbf{Non-sit} & \textbf{Sit} \\
\midrule
Non-sit & 0.77 & 0.23 \\
Sit & 0.12 & 0.88 \\
\bottomrule
\end{tabular}}
}
\end{minipage}
\vspace{12pt}
\begin{minipage}[t]{0.45\linewidth}
\centering
\subfloat[ViT\textsubscript{small}, Hip]{
\resizebox{0.7\linewidth}{!}{%
\begin{tabular}{l|cc}
 & \multicolumn{2}{c}{\textbf{Predicted}} \\
\cmidrule(lr){2-3}
\textbf{Actual} \quad & \textbf{Non-sit} & \textbf{Sit} \\
\midrule
Non-sit & 0.79 & 0.21 \\
Sit & 0.09 & 0.91 \\
\bottomrule
\end{tabular}}
}
\end{minipage}
\hfill
\begin{minipage}[t]{0.45\linewidth}
\centering
\subfloat[ViT\textsubscript{small}, Wrist]{
\resizebox{0.7\linewidth}{!}{%
\begin{tabular}{l|cc}
 & \multicolumn{2}{c}{\textbf{Predicted}} \\
\cmidrule(lr){2-3}
\textbf{Actual} \quad & \textbf{Non-sit} & \textbf{Sit} \\
\midrule
Non-sit & 0.69 & 0.31 \\
Sit & 0.11 & 0.89 \\
\bottomrule
\end{tabular}}
}
\end{minipage}

\caption{Confusion matrices for sitting versus non-sitting across 10-second windows. Top row: CHAP\textsubscript{FT}; bottom row: ViT\textsubscript{Small}. Left: hip; right: wrist. Each entry represents the fraction of samples for a given actual class that are predicted as the corresponding class.}
\label{tab:confusion_matrix_side_by_side}
\end{table}
\subsection{Subject-level Performance}
We further examine subject-level performance to quantify participant heterogeneity and localize the errors observed above. Metrics were computed per subject and then averaged across subjects. We report sensitivity (proportion of true sitting correctly classified), specificity (proportion of true non-sitting correctly classified), positive predictive value (PPV), negative predictive value (NPV), balanced accuracy (mean of sensitivity and specificity), and F1 score (harmonic mean of sensitivity and PPV). True positives are defined as correctly identified sitting. Results are presented in Table~\ref{tab: performance metrics}.

For the hip placement, CHAP\textsubscript{ZS} and CHAP\textsubscript{FT} performed nearly identically (balanced accuracy 0.87 vs.\ 0.86; F1 0.91 vs.\ 0.91), indicating that finetuning provides limited additional benefit for hip data. ViT\textsubscript{Small} was consistently lower across all metrics.

For the wrist placement, CHAP\textsubscript{FT} clearly outperformed both CHAP\textsubscript{ZS} and ViT\textsubscript{Small}. CHAP\textsubscript{FT} achieved the highest mean balanced accuracy (0.81) versus CHAP\textsubscript{ZS} (0.70) and ViT\textsubscript{Small} (0.79). The largest gap was in sensitivity (CHAP\textsubscript{FT}: 0.88 vs.\ CHAP\textsubscript{ZS}: 0.68). Differences between CHAP\textsubscript{FT} and ViT\textsubscript{Small} were much smaller ($ \leq 3$ \%), except for specificity and F1, where CHAP\textsubscript{FT} exceeded ViT\textsubscript{Small} by approximately 7\%.


Finally, we evaluated sedentary behavior accumulation patterns for each subject by computing four measures: (1) Total Sedentary Time (daily average minutes sedentary), (2) Sit-to-stand Transitions (daily average count), (3) Mean Sedentary Bout Duration (total sedentary time divided by total transitions), and (4) Minutes in Sedentary Bouts $\geq$30 min. These were computed for each model (CHAP\textsubscript{ZS}, CHAP\textsubscript{FT}, ViT\textsubscript{Small}) and for traditional intensity-based cut-point methods: ActiGraph 100 cpm \cite{Mathews} (hip vertical axis counts per minute $< 100$) and ActiGraph 1853 cpm \cite{Kuster} (wrist vector magnitude counts per minute $< 1853$).

\begin{table}[t]
\centering
\caption{Test set performance metrics across participants.}
\label{tab: performance metrics}
\setlength\tabcolsep{1pt}
\resizebox{\linewidth}{!}{%
\begin{tabular}{llcc cc cc}
\toprule
\textbf{Loc.} & \textbf{Metric} 
& \multicolumn{2}{c}{CHAP\textsubscript{ZS}} 
& \multicolumn{2}{c}{CHAP\textsubscript{FT}} 
& \multicolumn{2}{c}{ViT\textsubscript{Small}} \\
\cmidrule(lr){3-4}\cmidrule(lr){5-6}\cmidrule(lr){7-8}
& & \textbf{Mean(SD)} & \textbf{Median(IQR)} 
  & \textbf{Mean(SD)} & \textbf{Median(IQR)} 
  & \textbf{Mean(SD)} & \textbf{Median(IQR)}  \\
\midrule

\multirow{6}{*}{Hip} 
& Bal acc     
& 0.87(0.14) & 0.91 (0.06)
& 0.86 (0.13) & 0.91 (0.05)
& 0.83 (0.12) & 0.88 (0.09) \\
& Sensitivity 
& 0.90 (0.09) & 0.94 (0.06)
& 0.91 (0.08) & 0.94 (0.07)
& 0.90 (0.07) & 0.91 (0.06) \\
& Specificity 
& 0.83 (0.22) & 0.91 (0.10)
& 0.82 (0.22) & 0.90 (0.12)
& 0.77 (0.21) & 0.83 (0.13) \\
& PPV         
& 0.93 (0.14) & 0.97 (0.05)
& 0.93 (0.14) & 0.96 (0.06)
& 0.91 (0.13) & 0.94 (0.09) \\
& NPV         
& 0.72 (0.23) & 0.79 (0.22)
& 0.74 (0.23) & 0.81 (0.20)
& 0.71 (0.24) & 0.77 (0.19) \\
& F1          
& 0.91 (0.12) & 0.95 (0.05)
& 0.91 (0.12) & 0.95 (0.05)
& 0.90 (0.11) & 0.93 (0.05) \\
\midrule

\multirow{6}{*}{Wrist}
& Bal acc     
& 0.70 (0.12) & 0.72 (0.10)
& 0.81 (0.13) & 0.87 (0.10)
& 0.79 (0.13) & 0.83 (0.12) \\
& Sensitivity 
& 0.68 (0.08) & 0.90 (0.08)
& 0.88 (0.08) & 0.90 (0.08)
& 0.89 (0.06) & 0.92 (0.08) \\
& Specificity 
& 0.51 (0.23) & 0.83 (0.16)
& 0.75 (0.23) & 0.85 (0.16)
& 0.68 (0.23) & 0.75 (0.26) \\
& PPV         
& 0.88 (0.15) & 0.93 (0.12)
& 0.90 (0.15) & 0.93 (0.09)
& 0.87 (0.14) & 0.90 (0.09) \\
& NPV         
& 0.43 (0.20) & 0.48 (0.25)
& 0.68 (0.21) & 0.74 (0.21)
& 0.68 (0.23) & 0.74 (0.14) \\
& F1          
& 0.71 (0.13) & 0.79 (0.08)
& 0.80 (0.12) & 0.91 (0.07)
& 0.87 (0.12) & 0.90 (0.07) \\
\bottomrule
\end{tabular}%
}
\end{table}

Table~\ref{tab:agreementh} summarizes agreement between hip-based methods and SenseCam ground truth for subject-level sedentary pattern variables. In terms of mean absolute percentage error (MAPE), ViT\textsubscript{Small} performed best for total sedentary time (11.4\%) and time in sedentary bouts $\geq$30 min/day (15.5\%). For sit-to-stand transitions/day and mean bout duration, CHAP\textsubscript{FT} yielded the lowest errors (26.8\% and 29.2\%, respectively), with CHAP\textsubscript{ZS} showing comparable performance (27.7\% and 30.4\%). The ActiGraph 100 cpm method had substantially higher MAPEs for sit-to-stand transitions (246.7\%), time in bouts $\geq$30 min/day (53.5\%), and mean bout duration (70.7\%). Across all metrics, CHAP\textsubscript{FT} demonstrated consistently higher rank-based agreement with SenseCam than ViT\textsubscript{Small}, with Spearman correlations of 0.98, 0.85, 0.94, and 0.78, compared with 0.98, 0.79, 0.93, and 0.75 for ViT\textsubscript{Small}.

Table~\ref{tab:agreementw} presents agreement results for wrist-based estimates. Both CHAP\textsubscript{FT} and ViT\textsubscript{Small} exhibited substantially lower bias and error than CHAP\textsubscript{ZS} and the ActiGraph 1853 cpm cut point for wrist-based subject-level sedentary metrics. ViT\textsubscript{Small} showed the best overall performance, yielding the lowest mean absolute percentage error (MAPE) for total sedentary time (10.9\%), sit-to-stand transitions/day (31.5\%), time in sedentary bouts $\geq$30 min/day (15.0\%), and mean bout duration (27.1\%). CHAP\textsubscript{FT} generally showed the next lowest errors, although its MAPEs were consistently higher than those of ViT\textsubscript{Small}. Despite this, Spearman correlations for CHAP\textsubscript{FT} and ViT\textsubscript{Small} were broadly similar across the four metrics, indicating comparable ability to preserve the relative ranking of participants.
\begin{table*}[t]   
\centering
\caption{Agreement of hip-worn CHAP\textsubscript{ZS}, CHAP\textsubscript{FT}, ViT\textsubscript{Small}, and ActiGraph 100 cpm cut point with SenseCam for sedentary behavior metrics across participants}
\label{tab:agreementh}
\setlength\tabcolsep{6pt}
\resizebox{0.7\linewidth}{!}{%
\begin{tabular}{lp{2cm} p{2cm} p{2cm} p{2cm}}
\toprule
\textbf{Variable} & 
\textbf{Total sedentary time (min/day)} & 
\textbf{Sit-to-stand transitions/day} & 
\textbf{Time in bouts $\geq$30 min (min/day)} & 
\textbf{Mean bout duration (min)} \\
\midrule

\multicolumn{5}{l}{\textbf{Descriptive statistics}} \\ 
SenseCam, Mean (SD) & 318.68 (128.00) & 13.78 (5.42) & 220.58 (110.33) & 24.67 (9.81) \\
CHAP\textsubscript{ZS}, Mean (SD) & 281.42 (115.55) & 17.03 (6.22) & 169.92 (97.16) & 17.42 (6.46) \\
CHAP\textsubscript{FT}, Mean (SD) & 285.03 (117.80) & 16.92 (5.76) & 184.55 (104.26) & 17.58 (6.51) \\
CHAP-ViT, Mean (SD) & 291.53 (116.60) & 19.90 (7.26) & 203.03 (108.71) & 15.62 (6.44) \\
ActiGraph 100 cpm, Mean (SD) & 330.80 (126.96) & 47.77 (17.27) & 103.96 (82.04) & 7.22 (2.57) \\

\midrule
\multicolumn{5}{l}{\textbf{Agreement between CHAP\textsubscript{ZS} and SenseCam}} \\
Bias, Mean (SD) & $-37.26$ (32.78) & 3.24 (3.93) & $-50.66$ (38.09) & $-7.25$ (7.29) \\
Mean absolute error & 43.09 & 3.82 & 53.05 & 7.49 \\
Mean absolute percent error & 13.5\% & 27.7\% & 24.1\% & 30.4\% \\
Spearman correlation & 0.98 & 0.92 & 0.78 & 0.74 \\
Concordance correlation & 0.92 & 0.83 & 0.67 & 0.44 \\

\midrule
\multicolumn{5}{l}{\textbf{Agreement between CHAP\textsubscript{FT} and SenseCam}} \\
Bias, Mean (SD) & $-33.65$ (32.86) & 3.14 (3.19) & $-36.03$ (36.47) & $-7.10$ (7.10) \\
Mean absolute error & 41.18 & 3.69 & 40.56 & 7.21 \\
Mean absolute percent error & 12.9\% & 26.8\% & 18.4\% & 29.2\% \\
Spearman correlation & 0.98 & 0.85 & 0.94 & 0.78 \\
Concordance correlation & 0.93 & 0.72 & 0.89 & 0.46 \\

\midrule
\multicolumn{5}{l}{\textbf{Agreement between ViT and SenseCam}} \\
Bias, Mean (SD) & $-27.15$ (32.26) & 6.12 (4.85) & $-17.56$ (38.85) & $-9.05$ (8.14) \\
Mean absolute error & 36.32 & 6.32 & 34.10 & 9.32 \\
Mean absolute percent error & 11.4\% & 45.9\% & 15.5\% & 37.8\% \\
Spearman correlation & 0.98 & 0.79 & 0.93 & 0.75 \\
Concordance correlation & 0.94 & 0.49 & 0.92 & 0.32 \\

\midrule
\multicolumn{5}{l}{\textbf{Agreement between ActiGraph 100 cpm and SenseCam}} \\
Bias, Mean (SD) & 12.12 (47.55) & 33.99 (13.23) & $-116.62$ (59.39) & $-17.45$ (8.10) \\
Mean absolute error & 37.78 & 33.99 & 117.95 & 17.45 \\
Mean absolute percent error & 11.9\% & 246.7\% & 53.5\% & 70.7\% \\
Spearman correlation & 0.95 & 0.88 & 0.91 & 0.77 \\
Concordance correlation & 0.93 & 0.10 & 0.47 & 0.09 \\
\bottomrule
\end{tabular}%
}
\end{table*}

\begin{table*}[t]    
\centering
\caption{Agreement of wrist-worn CHAP\textsubscript{ZS}, CHAP\textsubscript{FT}, ViT\textsubscript{Small}, and ActiGraph 1853 cpm cut point with SenseCam for sedentary behavior metrics  across participants}
\label{tab:agreementw}
\setlength\tabcolsep{6pt}
\resizebox{0.7\linewidth}{!}{%
\begin{tabular}{l p{2cm} p{2cm} p{2cm} p{2cm}}
\toprule
\textbf{Variable} & 
\textbf{Total sedentary time (min/day)} & 
\textbf{Sit-to-stand transitions/day} & 
\textbf{Time in bouts $\geq$30 min (min/day)} & 
\textbf{Mean bout duration (min)} \\ 
\midrule

\multicolumn{5}{l}{\textbf{Descriptive statistics}} \\ 
SenseCam, Mean (SD) & 338.33 (128.41) & 13.93 (5.27) & 239.99 (116.20) & 25.87 (10.69) \\
CHAP\textsubscript{ZS} , Mean (SD) & 220.71 (104.75) & 83.03 (31.61) & 39.45 (47.08) & 2.72 (1.12) \\
CHAP\textsubscript{FT}, Mean (SD) & 310.46 (120.84) & 22.45 (9.02) & 207.76 (111.67) & 15.41 (8.54) \\
ViT, Mean (SD) & 322.64 (120.81) & 17.12 (7.00) & 241.12 (113.28) & 20.66 (9.55) \\
ActiGraph 1853 cpm, Mean (SD) & 300.62 (121.12) & 51.91 (18.36) & 73.79 (71.79) & 5.96 (2.31) \\

\midrule
\multicolumn{5}{l}{\textbf{Agreement between CHAP\textsubscript{ZS}  and SenseCam}}\\
Bias, Mean (SD) & $-117.62$ (67.43) & 69.10 (28.54) & $-200.53$ (97.48) & $-23.15$ (10.21) \\
Mean absolute error & 120.26 & 69.10 & 201.15 & 23.15 \\
Mean absolute percent error & 35.5\% & 496.1\% & 83.8\% & 89.5\% \\
Spearman correlation & 0.87 & 0.64 & 0.60 & 0.48 \\
Concordance correlation & 0.55 & 0.04 & 0.11 & 0.02 \\

\midrule
\multicolumn{5}{l}{\textbf{Agreement between CHAP\textsubscript{FT} and SenseCam}} \\
Bias, Mean (SD) & $-27.88$ (46.69) & 8.52 (6.13) & $-32.23$ (51.33) & $-10.46$ (7.24) \\
Mean absolute error & 48.14 & 8.54 & 49.35 & 11.00 \\
Mean absolute percent error & 14.2\% & 61.3\% & 20.6\% & 42.5\% \\
Spearman correlation & 0.92 & 0.81 & 0.87 & 0.77 \\
Concordance correlation & 0.91 & 0.39 & 0.86 & 0.45 \\

\midrule
\multicolumn{5}{l}{\textbf{Agreement between ViT and SenseCam}} \\
Bias, Mean (SD) & $-15.69$ (43.37) & 3.18 (4.50) & 1.13 (46.98) & $-5.21$ (7.38) \\
Mean absolute error & 36.74 & 4.39 & 35.89 & 7.02 \\
Mean absolute percent error & 10.9\% & 31.5\% & 15.0\% & 27.1\% \\
Spearman correlation & 0.93 & 0.78 & 0.90 & 0.71 \\
Concordance correlation & 0.93 & 0.65 & 0.92 & 0.65 \\

\midrule
\multicolumn{5}{l}{\textbf{Agreement between ActiGraph 1853 cpm and SenseCam}} \\
Bias, Mean (SD) & $-37.71$ (55.07) & 37.98 (14.48) & $-166.20$ (83.33) & $-19.90$ (9.17) \\
Mean absolute error & 53.10 & 37.98 & 166.84 & 19.90 \\
Mean absolute percent error & 15.7\% & 272.6\% & 69.5\% & 76.9\% \\
Spearman correlation & 0.88 & 0.81 & 0.66 & 0.62 \\
Concordance correlation & 0.86 & 0.08 & 0.25 & 0.07 \\
\bottomrule
\end{tabular}%
}
\end{table*}

\section{Sensitivity Analysis} \label{ablation}

\subsection{Weight Transfer vs. Reinitialization} 
We compared finetuning from pretrained CHAP weights \cite{Chap1.0} with training the same architecture from random initialization. Random initialization requires the model to learn accelerometer features from scratch using only iWatch data, whereas pretrained weights provide a starting point that already captures common signal statistics such as temporal patterns and cross-channel correlations.

\begin{figure}
    \centering
    \includegraphics[width=1\linewidth]{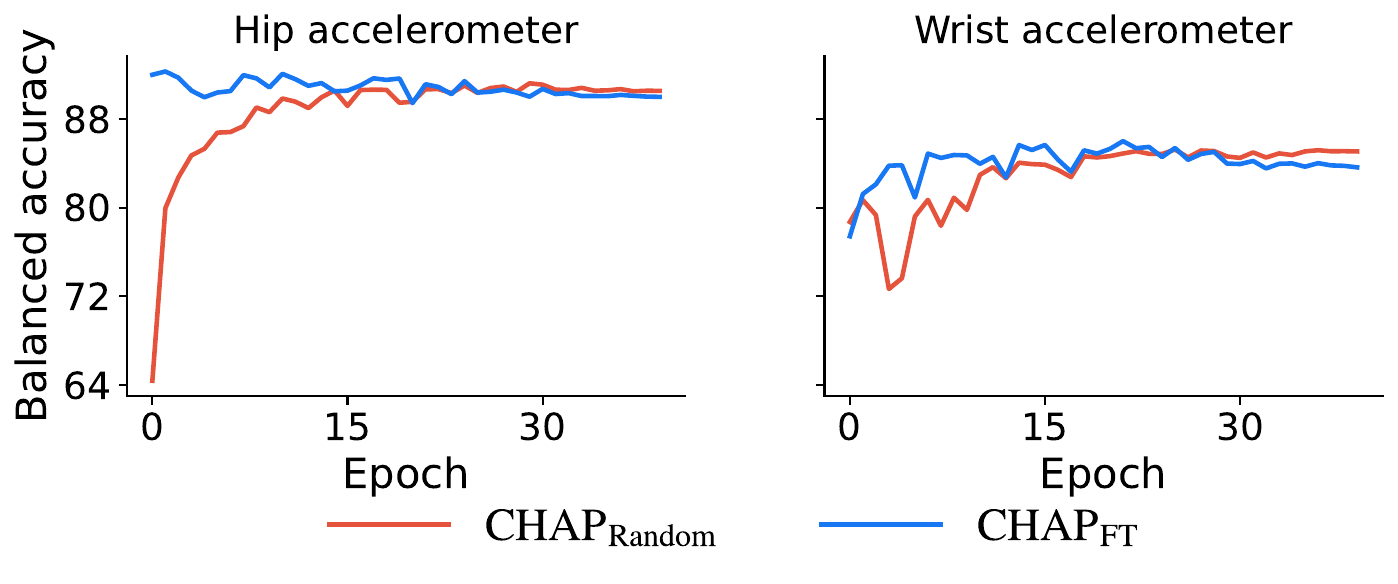}
    \caption{\textbf{Validation balanced accuracy over training epochs for CHAP models with random initialization and finetuning, evaluated on hip (left) and wrist (right) accelerometer data.} Finetuning improves early-epoch stability and convergence, especially for the wrist setting, while both models reach similar balanced accuracy after sufficient training.}
    \label{fig:chap_init}
\end{figure}

As shown in Figure~\ref{fig:chap_init}, CHAP\textsubscript{FT} reaches strong validation performance on iWatch hip almost immediately, whereas CHAP\textsubscript{Random} requires about 20 epochs to reach a similar level, consistent with the strong zero-shot hip performance in Table~\ref{tab:bal_f1_results}. For the wrist placement, CHAP\textsubscript{FT} starts higher during the first 10 epochs, but CHAP\textsubscript{Random} gradually closes the gap by around 20 epochs. This suggests that hip-pretrained weights provide a useful head start for wrist but the benefit diminishes with sufficient training.

\begin{table}
\centering
\caption{Balanced accuracy (Bal acc) and F1-score (\%) of CHAP\textsubscript{Random} Init and CHAP\textsubscript{FT} across hip and wrist placements.}
\label{tab:bal_f1_results}
\setlength\tabcolsep{6pt}
\resizebox{\linewidth}{!}{%
\begin{tabular}{llccc ccc}
\toprule
\textbf{Placement} & \textbf{Metric} 
& \multicolumn{3}{c}{CHAP Random Init.} 
& \multicolumn{3}{c}{CHAP\textsubscript{FT}} \\
\cmidrule(lr){3-5}\cmidrule(lr){6-8}
& & \textbf{Train} & \textbf{Validation} & \textbf{Test}
  & \textbf{Train} & \textbf{Validation} & \textbf{Test} \\
\midrule

\multirow{2}{*}{Hip} 
& Bal acc & 89.86 & 91.22 & 88.38  & 90.36 & 92.29 & \textbf{88.71} \\
& F1      & 91.56 & 92.98 & \textbf{90.35}  & 91.28 & 92.99 & 90.04 \\
\midrule

\multirow{2}{*}{Wrist} 
& Bal acc & 86.07 & 85.25 & 82.30  & 88.71 & 86.01 & \textbf{82.56} \\
& F1      & 87.67 & 87.80 & \textbf{85.09}  & 89.26 & 88.02 & 84.97 \\
\bottomrule
\end{tabular}%
}
\end{table}

\subsection{Limited Labels} 
\begin{figure}[hbt]
    \centering
    \includegraphics[width=1\linewidth]{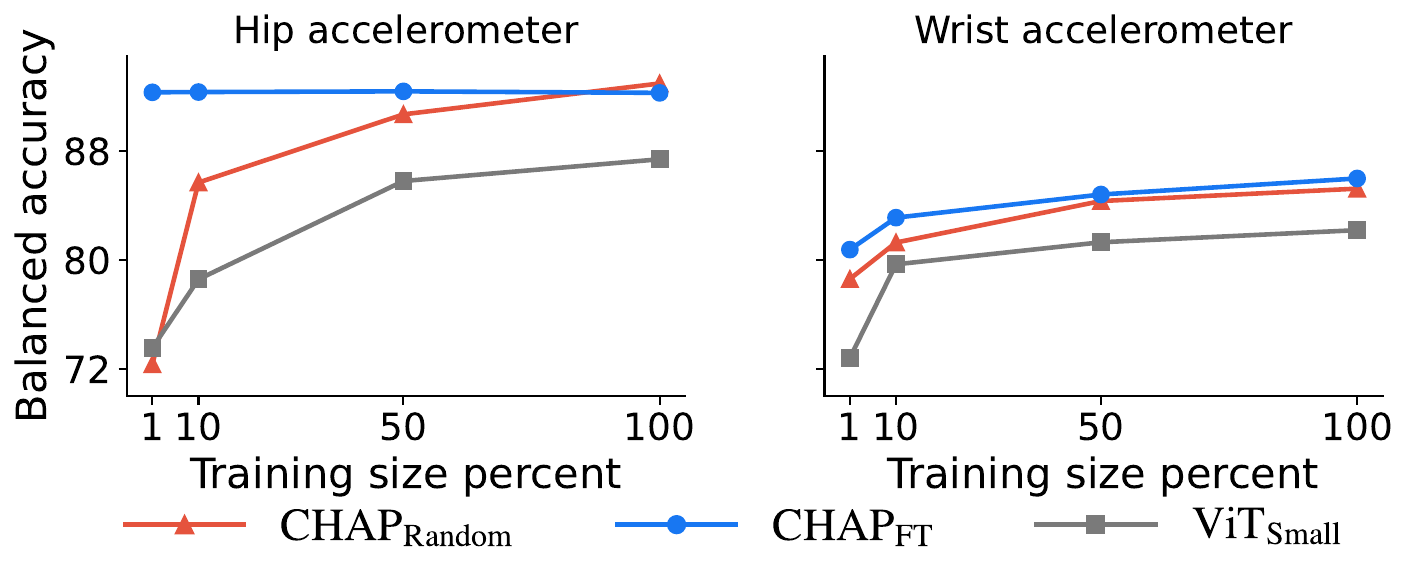}
    \caption{\textbf{Limited label analysis on (a) iWatch wrist and (b) hip placements.} Validation accuracy is reported as a function of labeled training size 1\%, 10\%, 50\%, and 100\% for CHAP\textsubscript{FT}, ViT\textsubscript{Small}, and CHAP\textsubscript{Random}.}
    \label{fig:limited_label}
\end{figure}
We further evaluated CHAP\textsubscript{Random}, CHAP\textsubscript{FT}, and ViT\textsubscript{Small} under limited labeled data by subsampling the training set to 1\%, 10\%, 50\%, and 100\% for both hip and wrist. Figure~\ref{fig:limited_label} summarizes validation accuracy trends and Table~\ref{tab:rf_results} reports full numerical results. This analysis targets a practical question: \textbf{how many new labels are needed for finetuning to help, and does accuracy saturate early enough to reduce annotation costs?}

For the \textit{hip placement} (Figure~\ref{fig:limited_label}a), CHAP\textsubscript{FT} is robust across all label budgets. CHAP\textsubscript{Random} is highly sensitive: while it matches CHAP\textsubscript{FT} at 100\% labels, performance drops sharply to 85.70\% at 10\% and 72.34\% at 1\%, illustrating that without pretrained weights the model overfits in low-data regimes. ViT\textsubscript{Small} shows similarly poor data efficiency (73.55\% and 78.59\% at 1\% and 10\% labels, respectively).

For the \textit{wrist placement} (Figure~\ref{fig:limited_label}b), the advantage of pretraining is markedly diminished. CHAP\textsubscript{FT} scales from 80.77\% at 1\% labels to 86.01\% at 100\%, but the gap over CHAP\textsubscript{Random} remains narrow throughout, consistent with a hip-to-wrist domain mismatch in motion dynamics (Figure~\ref{fig:joint_hexbin_hip_wrist}). Wrist accuracy remains consistently lower than hip accuracy regardless of label budget. This ceiling likely reflects higher signal variability at the wrist: common upper-limb activities (i.e., writing, typing) occur during sedentary periods, introducing intra-class noise that makes sitting harder to isolate than with the more stable trunk motions captured at the hip.

ViT\textsubscript{Small} trained from scratch underperformed CHAP on both placements, even relative to CHAP\textsubscript{Random}, despite having more parameters. This likely reflects differences in inductive bias: CHAP encodes temporal structure through bidirectional LSTMs and local feature extraction, providing priors well suited to our setting where labels are temporally smooth and transitions are rare. A transformer relies primarily on attention with weaker assumptions about local temporal dynamics, making performance more sensitive to training set size \cite{ViT,scaling}. Future work could explore transformer pretraining on larger unlabeled accelerometer corpora to reduce this gap.
\begin{table}
\centering
\caption{\textbf{Model performance on hip and wrist placements when varying training size (\%).} Results are reported as validation and test accuracy.}
\label{tab:rf_results}
\resizebox{\linewidth}{!}{%
\setlength{\tabcolsep}{4pt}
\begin{tabular}{llcc cc cc}
\toprule
& & \multicolumn{2}{c}{CHAP Random Init} & \multicolumn{2}{c}{CHAP\textsubscript{FT}} & \multicolumn{2}{c}{ViT\textsubscript{Small}} \\
\cmidrule(lr){3-4}\cmidrule(lr){5-6}\cmidrule(lr){7-8}
\textbf{Placement} & \textbf{Training size (\%)} 
& \textbf{Valid acc.} & \textbf{Test acc.} 
& \textbf{Valid acc.} & \textbf{Test acc.} 
& \textbf{Valid acc.} & \textbf{Test acc.} \\
\midrule
\multirow{4}{*}{Hip}
 & 1   & 72.34 & 68.80 & 92.33 & 88.73 & 73.55 & 70.62 \\
 & 10  & 85.70 & 82.27 & 92.35 & 88.76 & 78.59 & 76.24 \\
 & 50  & 90.71 & 87.74 & 92.40 & 88.65 & 85.81 & 84.11 \\
 & 100 & 92.98 & 90.35 & 92.29 & 88.71 & 87.41 & 85.22 \\
\midrule
\multirow{4}{*}{Wrist}
 & 1   & 78.61 & 76.88 & 80.77 & 78.53 & 72.81 & 71.16 \\
 & 10  & 81.29 & 77.91 & 83.12 & 80.33 & 79.69 & 77.40 \\
 & 50  & 84.34 & 80.90 & 84.82 & 81.74 & 81.31 & 78.80 \\
 & 100 & 85.25 & 82.30 & 86.01 & 82.56 & 82.20 & 79.39 \\
\bottomrule
\end{tabular}%
}
\end{table}


\section{Visualization}
\begin{figure}[!t]
    \centering
    \resizebox{\linewidth}{!}{%
        \includegraphics[trim=1cm 12cm 1cm 2cm,clip]{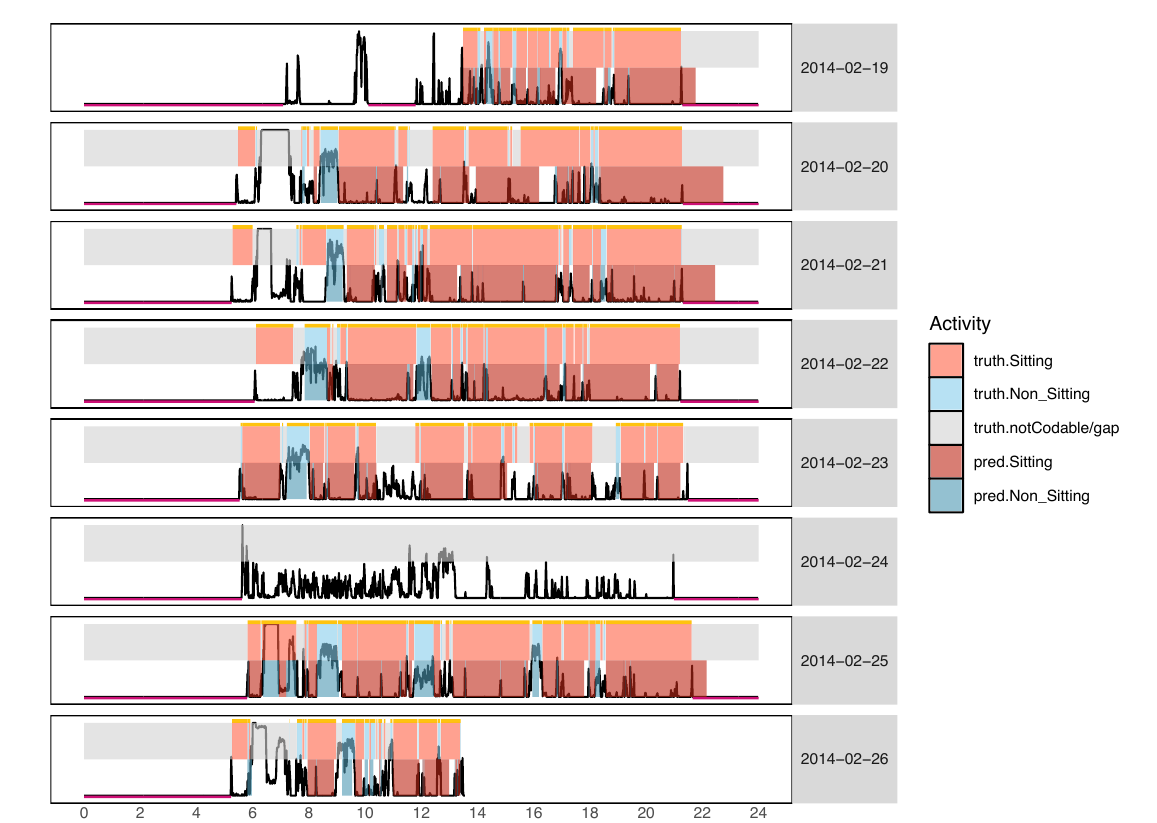}%
    }
    \vspace{2pt}
    \resizebox{\linewidth}{!}{%
        \includegraphics[trim=1cm 12cm 1cm 2cm,clip]{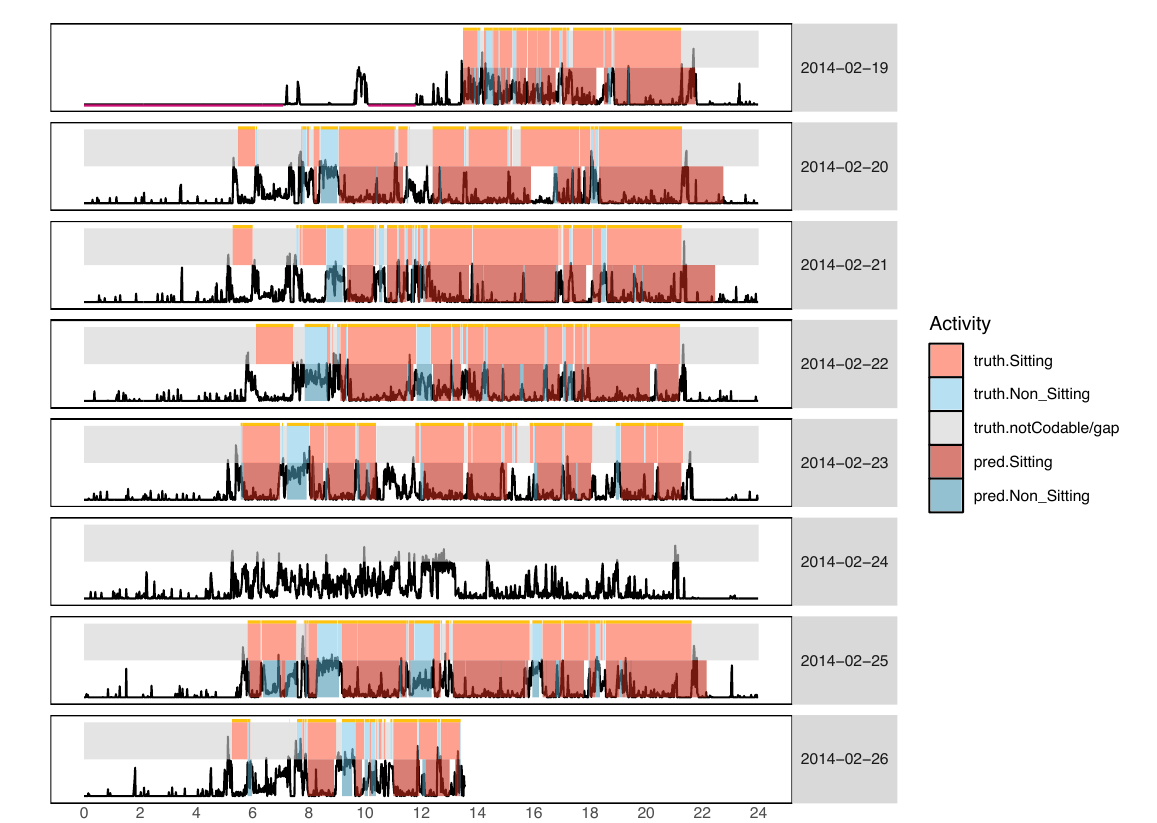}%
    }
    \vspace{-6pt}
    \caption{\textbf{CHAP\textsubscript{FT} prediction heatmaps for subject \textbf{i067A} with hip placement (top figure) and wrist placement (bottom figure).}
    \textbf{Yellow band:} wear log.
    \textbf{Hot Pink band:} non wear segments (extracted from ActiGraph via the Choi algorithm).
    \textbf{Black:} Vector Magnitude  per minute from ActiGraph.
    \textbf{Solid bars:} posture labels.
    Ground truth activities are shown in the \textbf{upper half}, and prediction results are shown in the \textbf{lower half}.}
    \vspace{-8pt}
    \label{fig:i0167A_heatmap_hw}
\end{figure}

Figure~\ref{fig:i0167A_heatmap_hw} shows CHAP\textsubscript{FT} prediction heatmaps for an illustrative subject (i0167A), where balanced accuracy exceeded 96\% for both placements. Predictions were generally consistent with SenseCam labels, with occasional mismatches during brief acceleration fluctuations.


\section{Implementation}
\label{implementation}
\textbf{Model Configurations.}
Table~\ref{tab:transformer_arch} provides the ViT\textsubscript{Small} architecture; CHAP details are in~\cite{Chap1.0}.
\begin{table}[h]
\centering
\caption{\textbf{Architecture of the transformer classifier.}}
\label{tab:transformer_arch}
\setlength{\tabcolsep}{3pt}
\renewcommand{\arraystretch}{0.95}
\resizebox{\linewidth}{!}{%
\begin{tabular}{l l c c}
\toprule
\textbf{Layer} & \textbf{Config} & \textbf{Output size} & \textbf{Depth} \\
\midrule
Input & -- & $42 \times 100 \times 3$ & -- \\
Patch embedding & MLP, embed dim 384 & $(42 \times 3) \times 384$ & 1 \\
Encoder blocks & MHSA + MLP & $(42 \times 3) \times 384$ & 12 \\
Encoder norm & LayerNorm & $(42 \times 3) \times 384$ & 1 \\
Mean pooling & Mean over tokens & $42 \times 384$ & 1 \\
Classifier & Linear & $42 \times 1$ & 1 \\
\bottomrule
\end{tabular}%
}
\end{table}

\textbf{Loss Function.} As noted in Section~\ref{data}, sitting occurs 2--3 times more often than non-sitting. We use a class-weighted binary cross-entropy loss with weight $w = n_{\text{sit}} / n_{\text{non-sit}}$ applied to the minority class, giving $w=2.80$ for hip and $w=2.82$ for wrist. This weighting improved CHAP\textsubscript{FT} validation balanced accuracy from 92.15\% to 92.29\% on hip and from 84.23\% to 86.01\% on wrist.

\textbf{Training Configuration.} Both CHAP\textsubscript{FT} and ViT\textsubscript{Small} are trained for 40 epochs with 8-epoch warmup, using AdamW (learning rate $1\mathrm{e}{-3}$, weight decay $1\mathrm{e}{-3}$), cosine learning rate schedule, and batch size 128.

\begin{figure*}[!t]
    \centering
    \includegraphics[width=1\linewidth]{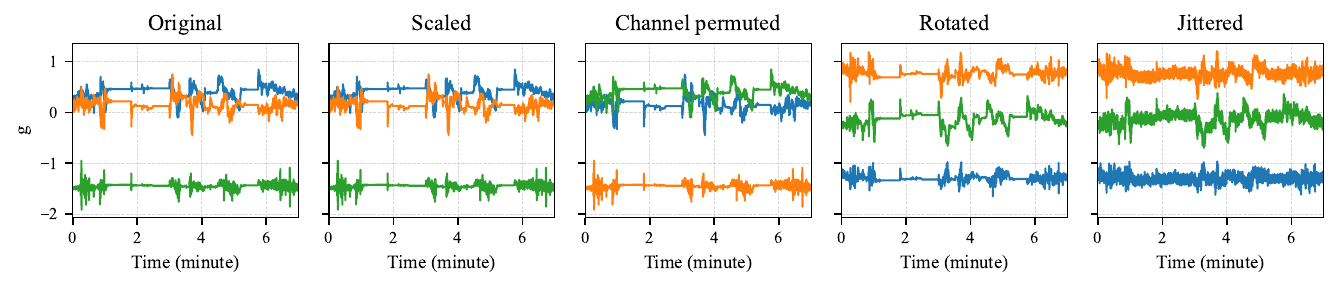}
    \caption{Data augmentation: a seven-minute sample of tri-axial accelerometer data is shown in its original form followed by four types of augmentations added step by step: scaling, channel permutation, axis rotation, and jittering.}
    \label{fig:augment_steps}
\end{figure*}

\textbf{Data Augmentation.} We applied jittering, scaling, channel permutation, and axis rotation during training~\cite{harnet} (Figure~\ref{fig:augment_steps}) to encourage invariance to sensor orientation and structural noise.

\section{Discussion}\label{sec12}

CHAP shows strong hip performance even in zero-shot settings, consistent with prior studies \cite{Chap1.0}.This finding shows that pretrained models can maintain accuracy when applied within the same body placement, which makes them valuable in low-label setings.

When the hip-trained model is applied directly to wrist sensors, accuracy drops significantly, but finetuning with as little as 10\% of labeled wrist data yields substantial improvement, establishing hip-based pretraining as a practical strategy for rapid adaptation. 

The performance gap between hip versus wrist placements could reflect
biomechanical
differences: the hip provides a more stable posture signal, while the wrist exhibits higher variability from arm movements unrelated to posture.
Despite this challenge, wrist placement remains important clinically because it is less intrusive, yields higher wear compliance, supports concurrent sleep assessment, and is already prevalent in commercial trackers such as Fitbits and Apple watches. These benefits make wrist-specific adaptation essential for translating sensor-based models into practice.

\section{Conclusion}\label{sec13}
In this work, we tested whether the hip-trained CNN-BiLSTM model (CHAP) can transfer to wrist accelerometer data for sitting versus non-sitting classification in the iWatch cohort. The hip model transfers well to unseen hip data without retraining, but wrist performance drops due to distributional differences. Finetuning on wrist data improves balanced accuracy and reduces the systematic error of predicting low-motion non-sitting as sitting. Under matched training conditions, finetuned CHAP is competitive with or better than a transformer trained from scratch. Sensitivity analyses show that hip-pretrained weights mainly benefit early training and data efficiency; with enough wrist labels and training time, random initialization reaches similar accuracy, highlighting the need for wrist-specific adaptation rather than assuming placement-agnostic features. Future work should explore pretraining with multiple placements and larger unlabeled wrist corpora, and consider complementary signals to reduce wrist ambiguity during posture inference in free-living settings.

\section*{ Conflict of Interest Statement}
Paul R. Hibbing received funding from Ametris LLC through a Digital Endpoint Accelerator Research grant. Other co-authors do not report any conflicts of interest. 

 \section*{Acknowledgment}
 The authors would like to thank the Cognitive Hardware and Software Ecosystem, Community Infrastructure (CHASE-CI) at the University of California San Diego for providing high performance computing cluster support (National Science Foundation grants 2100237 and 2120019). Also, we are grateful to the iWatch study team and participants who provided the data for this study. JZ's work was also supported by the National Heart, Lung, and Blood Institute of the National Institutes of Health Award R01HL166802.

\section*{References}
\bibliographystyle{IEEEtran}
\bibliography{bibliography}

\appendices

\end{document}